\documentclass[pmlr]{jmlr}
\usepackage{float}

\RequirePackage{graphicx}
 \usepackage{booktabs}
\usepackage{longtable}
\usepackage{booktabs} 
\usepackage{array}    
\usepackage{siunitx}  
\usepackage{caption}  

\makeatletter
\def\set@curr@file#1{\def\@curr@file{#1}} 
\makeatother
\usepackage[load-configurations=version-1]{siunitx} 
\usepackage{makecell}
\usepackage{multirow}
\usepackage{capt-of}

\theorembodyfont{\upshape}
\theoremheaderfont{\scshape}
\theorempostheader{:}
\theoremsep{\newline}

\jmlrvolume{252}
\jmlryear{2024}
\jmlrworkshop{Machine Learning for Healthcare}

\title[Deep Learning for Pathological Gait Classification]{Benchmarking Reliability of Deep Learning Models for Pathological Gait
Classification}

\author{\Name{Abhishek Jaiswal}
       \Email{abhi.jaiswal44@gmail.com}\\ 
       \addr Computer Science and Engineering\\
       Indian Institute of Technology Kanpur\\
       Kanpur, Uttar Pradesh, India 
       \AND
       \Name{Nisheeth Srivastava}
       \Email{nsrivast@iitk.ac.in}\\
        \addr Computer Science and Engineering\\
       Indian Institute of Technology Kanpur\\
       Kanpur, Uttar Pradesh, India }

\begin{document}

\maketitle

\begin{abstract}
Early detection of neurodegenerative disorders is an important open problem, since early diagnosis and treatment may yield a better prognosis. Researchers have recently sought to leverage advances in machine learning algorithms to detect symptoms of altered gait, possibly corresponding to the emergence of neurodegenerative etiologies. However, while several claims of positive and accurate detection have been made in the recent literature, using a variety of sensors and algorithms, solutions are far from being realized in practice. This paper analyzes existing approaches to identify gaps inhibiting translation. Using a set of experiments across three Kinect-simulated and one real Parkinson's patient datasets, we highlight possible sources of errors and generalization failures in these approaches. Based on these observations, we propose our strong baseline called Asynchronous Multi-Stream Graph Convolutional Network (AMS-GCN) that can reliably differentiate multiple categories of pathological gaits across datasets.
\end{abstract}

\section{Introduction}

\captionsetup{
  font=small, 
  labelfont=bf, 
  format=plain, 
  width=\textwidth, 
}
\newcolumntype{L}[1]{>{\raggedright\arraybackslash}p{#1}}
\newcolumntype{R}[1]{>{\raggedleft\arraybackslash}p{#1}}
\newcolumntype{C}[1]{>{\centering\arraybackslash}p{#1}}

Gait impairments, particularly those resulting from Parkinson's Disease(PD), Alzheimer's disease (AD), and other neurodegenerative disorders, cause muscle stiffness, dis-coordination,  tremors, and posture instability and severely reduce peoples' quality of life. Freezing of gait (FOG), visible in more severe conditions, significantly increases the risk of falls and fractures~\citep{coste2014detection}. Gait symptoms such as slowness also correlate with the onset of dementia \citep{mielke2013assessing}. Aging increases the risk of gait disorders, rising from 10 \% to more than 60 \% in people from 60 to 80 years of age \citep{mahlknecht2013prevalence}.  Despite this criticality, gait asymmetries and deficits largely remain underdiagnosed and receive subpar medical attention. \citep{rubenstein2004detection}. 

Early diagnosis of gait-related disorders reduces the overall health care cost and initiates effective therapy routine \citep{pistacchi2017gait}. As a possible method for reducing costs of such diagnosis at scale, automatic pathological gait classification has received significant interest in recent times, with most prominent approaches using customized sensor arrays for gait tracking~\citep{xu2021machine,schlachetzki2017wearable,um2017data,kubota2016machine}.  While highly accurate, sensor-based methods are inconvenient and obtrusive, thus limiting their usability for the general populace. Sensor arrays require accurate placement, rendering them implausible for self- or weakly-supervised diagnosis of gait problems. 

Non-invasive computer vision methods employing machine learning techniques turn out to be a promising replacement for such sensor-based gait analysis.
For instance, features from gait video sequences can be extracted to classify gait as healthy or abnormal~\citep{nieto2016vision,ajay2018pervasive,nguyen2016skeleton}. Similarly,
Silhouette-based feature extraction used in biometric technology, has possible extensions to pathological gait classification~\citep{albuquerque2021remote,ortells2018vision,loureiro2020using}. However, being appearance-based, such models collapse with scale and clothing variations~\citep{wang2010review}.

 Common gait anomalies like lurching, freezing, and stiff-legged gait can be imitated under supervision. As health data is costly, such simulations are invaluable.
They can be recorded through the use of knee and sole pads, combined with restricted joint movements. 
Recent studies show that an affordable Kinect can efficiently capture skeleton information~\citep{meng2016detection,li2018classification,jun2020feature,pachon2020abnormal}. Furthermore, the recorded data is considered precise enough to be suitable for training and classification~\citep{stone2011evaluation,clark2019three}.
  The Multi-Modal Gait Symmetry dataset (MMGS)~\citep{khokhlova2019normal}, Walking Gait dataset~\citep{nguyen20183d}, and Pathological Gait dataset~\citep{jun2020pathological} are three such Kinect-based datasets with simulated gait pathologies.
  
 Graph Networks are well suited to model structured information, with applications in pose estimation~\citep{xu2021graph}, motion modeling~\citep{huang2019stgat,mohamed2020social, jaiswal2023using} and physics prediction~\citep{battaglia2016interaction,sanchez2020learning}, among others.
 Of them, the Spatio-Temporal Graph Convolutional Networks (STGCN), specially designed for processing skeleton information, have been prolifically used for skeleton-based action recognition~\citep{yan2018spatial,cheng2020skeleton} and are currently state-of-the-art in the task of pathological gait classification.

 But as we discuss in detail below, several methods proposed for gait classification premise on individual models fitting to only a single dataset. However, we view this approach as a critical limitation. Robustness and fairness are paramount to healthcare applications, where inaccurate predictions could significantly affect the well-being of individuals and the effectiveness of treatment protocols. Therefore, addressing these limitations and exploring approaches that ensure reliable and accountable gait classification is crucial. This paper aims to tackle these challenges.
 \begin{figure*}[tb]
\centering
\includegraphics[width=\textwidth]{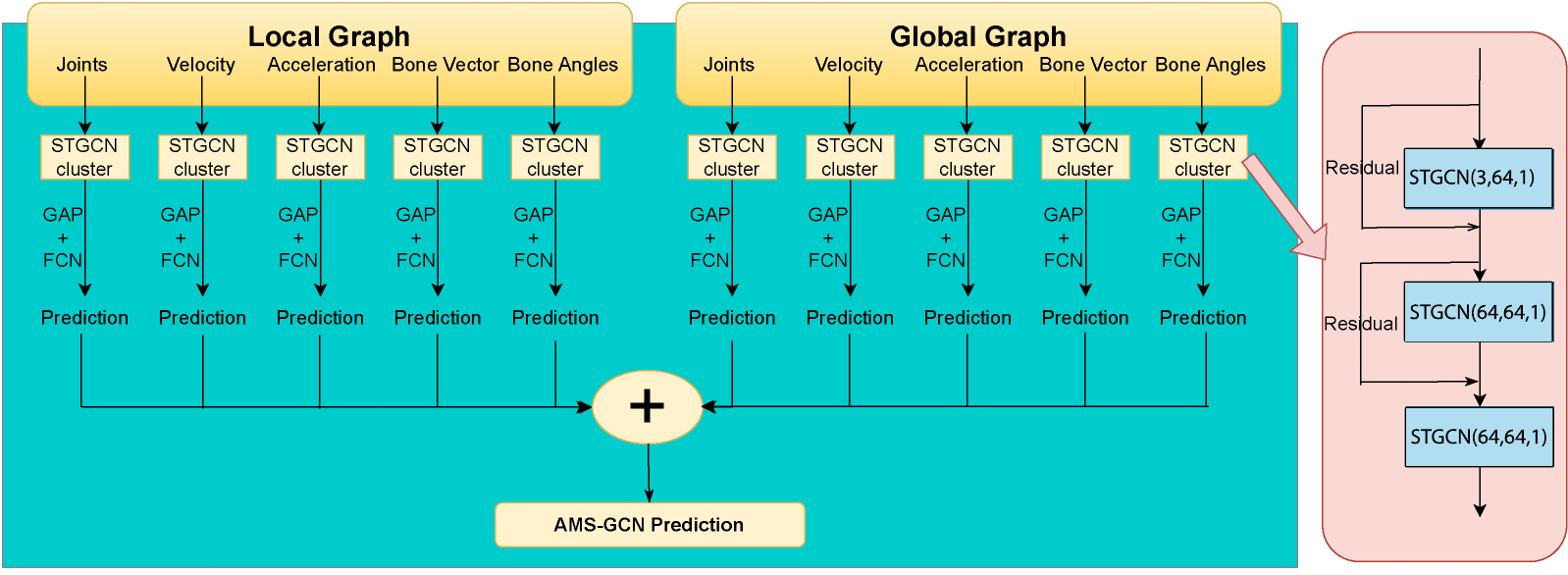}
\caption{Our proposed Expert Committees ensemble model. GAP = Global Average Pooling, FCN~=~Fully Connected Network. All individual model predictions are normalized and added to get final scores.}
\label{fig:AMSGCN}
\end{figure*}

\begin{figure}[tbp]
\centering
\includegraphics[width=0.5\linewidth]{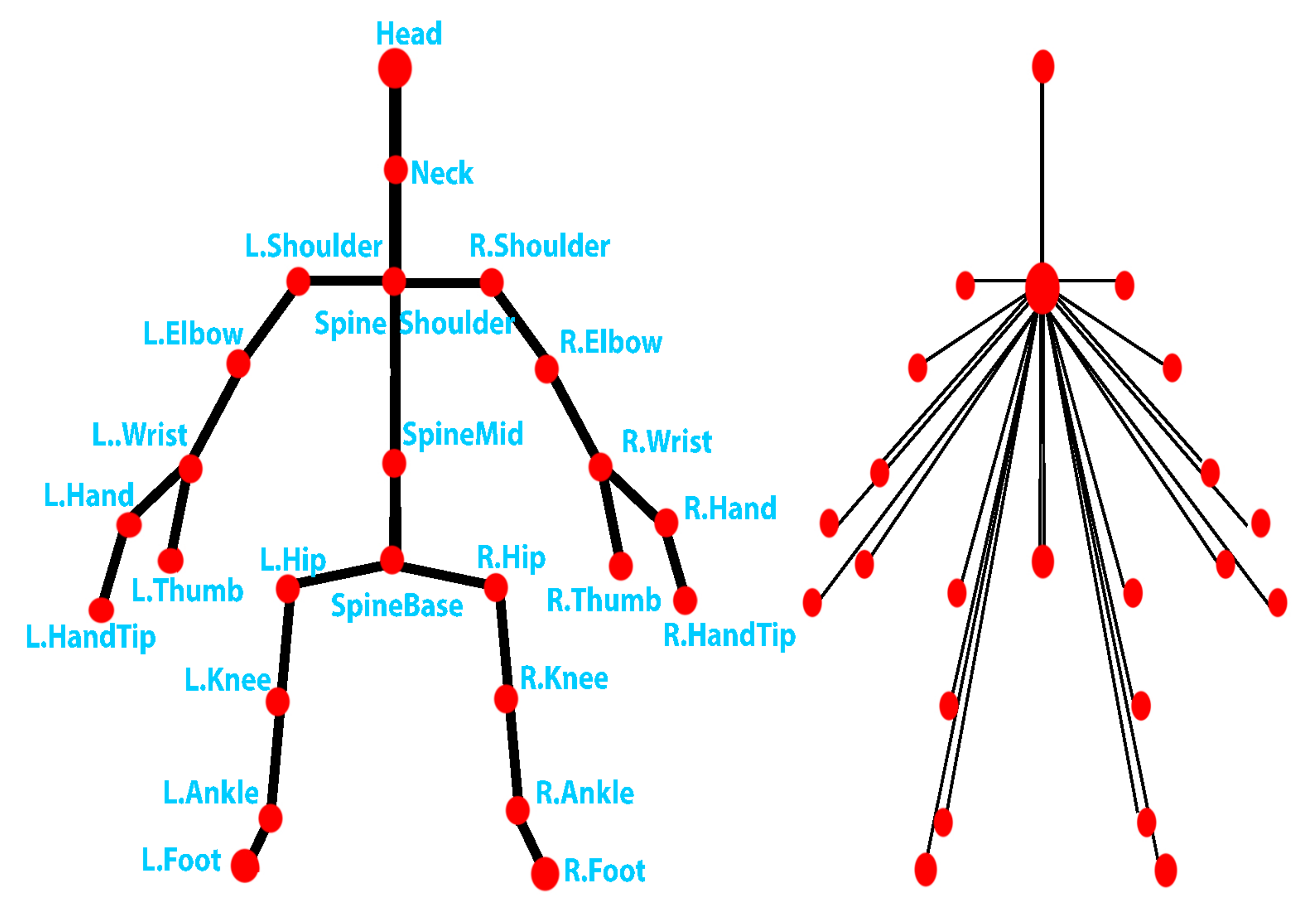}
\caption{Local(Left) and Global(Right) graphs along with 25 Joints of Kinect}
\label{fig:kinect25}
\end{figure}

Specifically, we aim to provide quantitative benchmarks for the ability of state-of-the-art pathological gait classification models. 
To this end, we conduct various experiments across three simulated pathological gait datasets and one dataset incorporating real gait pathologies (Section \ref{benchmark_study}).
Consequently, identifying large distribution shifts in the most acclaimed gait classification models, we also propose and evaluate a new baseline for pathological gait classification,  Asynchronous Multi-Stream Graph Convolution Network (AMS-GCN) (Figure~\ref{fig:AMSGCN}), demonstrating robust classification performance across datasets (Section \ref{amsgc_intro} and \ref{amsgcn_analysis}). AMS-GCN ensembles gait features over local and global graphs (Figure~\ref{fig:kinect25}), validating stand-alone feature efficiency and furnishing critical clinical insights. Moreover, it demonstrates nearly identical performance across both validation and test sets, distinguishing itself from the outcomes of previous methods. 

Additionally, we evaluate the AMS-GCN with various feature combinations and ensembling techniques, drawing out strategies effective for practical applications in healthcare (Section \ref{amsgcn_insights}).






\subsection*{Generalizable Insights about Machine Learning in the Context of Healthcare}
\begin{itemize}

\item We explore the generalizability of existing gait classification algorithms on multiple datasets, highlighting potential causes inhibiting test time translation. Consequently, we show that it is easy to overfit healthcare data tested in isolation, which is antithesis to robustness critical to healthcare.

\item We propose an Asymmetric Multi-Stream Graph Convolutional Network, holistically processing motion information to assist pathological gait identification in home settings. Here we emphasize that, in many instances, surprisingly simple combinations can outperform overly complicated, poorly tested models.

\item Finally, using various ablation studies, we draw insights into the relative importance of different gait features used for classification. This opens up avenues for complimenting existing geriatric evaluations with additional gait parameters, thereby supporting clinical studies.
\end{itemize}



\section{Gait Classification}
 Markerless vision systems like camera systems, even though unobtrusive, are prone to errors and produce noisy joint coordinates. 
 To validate the Kinect approach and provide a foundation for future research, \citet{stone2011evaluation} assessed the reliability of Kinect-based gait parameters, including walking speed, stride time, and stride length, against marker-based Vicon Motion Capture systems. Their findings identified these parameters as suitable for detecting early-onset of gait abnormalities and assessing fall risk.

Employing feature engineering to model spatiotemporal relationships, \citet{chaaraoui2015abnormal} utilized Joint Motion History to diagnose gait abnormalities, focusing on a small set of 56 data sequences.
For binary classification across nine gait classes, \citet{nguyen2016skeleton} trained an HMM model on body limb angles to diagnose anomalies using a log-likelihood threshold.

In a non-Kinect-based setting, \citet{ajay2018pervasive} classified Parkinsonian gait directly from YouTube videos recorded from pervasive devices like smartphones and webcams. Meanwhile, \citet{li2018classification} encoded 3D joint motion trajectories, computing joint correlations for classification on a private dataset. These approaches avoid using joints' spatiotemporal information directly to reduce noise but are only functional in straightforward scenarios.

Others have explored the use of sequential and convolutional end-to-end architectures to broaden the scope of results. For instance, \citet{khokhlova2019normal} employed 3D knee and hip flexion angles, along with rotation data extracted from Kinect gait cycles, to train an LSTM ensemble, effectively reducing result variance.

Given that gait impairments correspond to a decline in specific limb functionalities, \citet{jun2020pathological} developed a GRU-based classifier using 3D skeletal data, and they observed classification scores with different joint groups.
In another approach, \citet{sadeghzadehyazdi2021modeling} utilized a CNN-LSTM model with KL divergence loss on normalized joints data, applying transfer learning on a different simulated dataset.
Taking an encoding-based approach, \citet{jun2020feature} extracted gait features from a Recurrent Neural Network Autoencoder (RNN-AE) and compared these features with multiple discriminative models. However, these studies primarily present results on one or two datasets, necessitating further examination.

Recent research has increasingly turned towards Graph Networks for modeling human pose, with a specific emphasis on leveraging Spatio-Temporal Graph Convolutional Networks (STGCN) \citep{yan2018spatial} as the foundational element for gait classification.
For example, \citet{he2022integrated} utilized a dual-branched STGCN model to independently process global and local features, capturing subtle manifestations of Parkinson's disease. They recorded and validated their model using a dataset from Parkinson's patients (PD-Walk).
In another study, \citet{tian2022skeleton} implemented a three-way data partitioning scheme on seven lower limb joints, incorporating joint, bone, and sagittal symmetry features along with an attention mechanism.
Similarly, \citet{kim2022pathological} constructed a multi-branched bottleneck architecture with five STGCN blocks and attention modules, presenting results on publicly available activity recognition and gait datasets. 

However, these methods require further investigation before they can be put into practice, as multi-STGCN systems could be susceptible to overparameterization. As such, we undertake a more in-depth analysis of these results in subsequent sections. Additionally, we offer insights into the features and model specifications that prove effective and highlight errors made by previous works. This analysis forms the first part of our contributions.

In the latter part of the paper, leveraging our insights, we construct an expert committee model for diagnosing pathological gait, while also exploring various modalities of prediction amalgamation. Surprisingly, a simpler and more straightforward model emerges as the clear winner, establishing itself as a robust baseline for any work on pathological gait classification.

\section{Benchmarking Study}
\label{benchmark_study}
\subsection{Dataset Statistics}

 Video-based pathological gait detection can assist geriatricians and patients with quantitative gait measurements and support for early diagnosis, cutting down frequent hospital commutes, particularly for patients with mobility impairments. 

In the absence of real patient data, researchers have recognized simulated walks as a fitting differentiator between standard gait patterns and multifarious abnormalities. Here, we present a succinct overview of three prominent simulated datasets repeatedly used for model evaluations for pathological gait classification.

\textbf{Multi-Modal Gait Symmetry (MMGS) \citep{khokhlova2019normal}.}
This dataset contains three gait categories - normal, limping gait, and knee rigidity gait, with 27 subjects walking each category five to seven times. Limping is simulated by affixing a 7cm padding sole to the right shoe of the subject. The third gait is simulated by constraining knee flexion while walking.

\textbf{Pathological Gait~\citep{jun2020pathological}.}
This dataset contains five pathological gaits - antalgic, stiff-legged, lurching, steppage, and Trendelenburg gaits and one normal category, recorded via six Kinect devices calibrated to be mutually consistent. Each subject walked 20 times for each gait type, and thus this dataset contains 10 subjects x 6 gait types x 120 walks(6 Kinect x 20 times) = 7200 total data clips from 6 Kinects.

\textbf{Walking Gait~\citep{nguyen20183d}.}
The Walking Gait dataset is recorded via a treadmill and contains nine gait types where the eight abnormal gaits are simulated by padding a sole of 5, 10, and 15 cm under the left and right feet and by attaching a weight of 4-kilogram to the left and right ankles.  Nine subjects walk for each of these nine gait types, with each walk containing 1200 frames.

Apart from these, we compare models on \textbf{PD-Walk dataset}~\textbf{\citep{he2022integrated}} which contains data from actual Parkinson's Disease(PD) patients. This dataset has skeleton joints data from walking videos for 96 PD patients and 95 healthy subjects.

For more details on these datasets, the readers are requested to refer to the specific dataset papers.

While prior results in the gait classification literature appear optimistic, robust assessments of these proposals are not always available. For instance, evaluations often lack validation sets and fail to specify essential hyperparameter details.
To enhance accountability and ensure accurate reporting, results should be demonstrated on multiple train, validation, and test splits, and yet many papers are lax in reporting the data splits used.

Even though we have attempted to optimize all relevant hyperparameters in our benchmarking study, it is plausible that specific hyperparameters we could not discover might shift the classification scores closer to claimed values. Nevertheless, our study proceeds from the premise that models should be reproducible based on the details reported in publications, and we meticulously outline all essential implementation assumptions to uphold transparency.

\begin{figure}[tbp]
\centering
\includegraphics[width=0.6\linewidth]{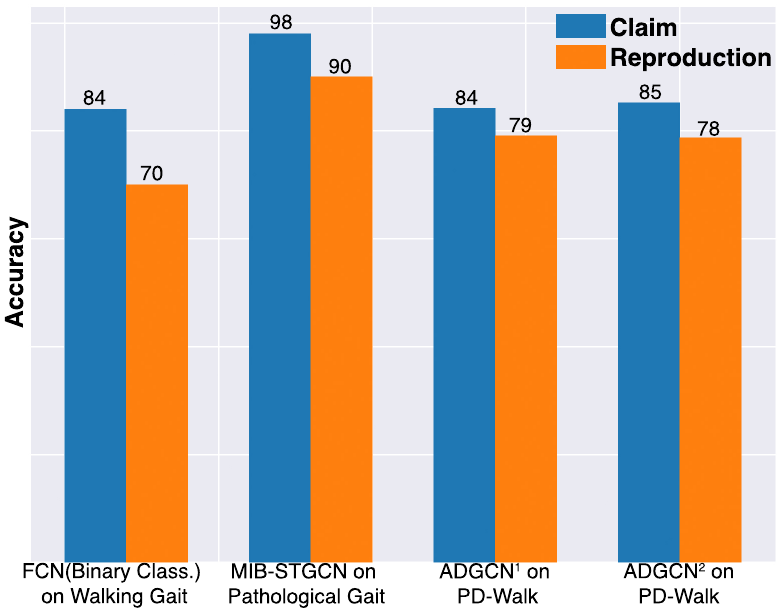}
\caption{Claimed and Reproduction Accuracies for existing models. FCN claimed results for binary classification on Walking Gait dataset use data preprocessing but no data augmentation. All MIB-STGCN results are without attention mechanism. 2s-AGCN has prominently been applied to Human Activity Recognition and has no reliable investigation for Gait Analysis.[1] - Reimplemented using Cross Entropy Loss. [2] - Reimplemented using Focal loss. All Reproduction results with multiple runs over multiple splits.}
\label{fig:distributionShift}
\end{figure}




\subsection{Contemporary Models}
For this study, we examine the best-performing gait classification models based on established machine learning architectures. To thoroughly cover the space of prior works, we focus on FCN (Fully Convolutional Networks), 2s-AGCN (Two-Stream Adaptive Graph Convolutional Networks), MIB-STGCN (Multiple-Input Branch STGCN), and ADGCN (Asymmetric Dual-Stream Graph Convolution Network), the last three being variants of the celebrated STGCNs coming from the skeleton action recognition setting. Below, we briefly describe the key distinctions among these models.

\textbf{\citet{sadeghzadehyazdi2021modeling}} evaluated multiple models on the Walking Gait dataset \citep{nguyen20183d}, employing eighteen normalized joints and KL divergence loss. We specifically focus on their \textbf{Fully Convolutional Network (FCN)} model, featuring 650 input-output channels in the convolutional layers, followed by a fully connected layer. We adopt their data preprocessing approach, normalizing data between -1 and 1, and apply our data augmentation technique for consistent evaluation (refer Appendix Section \ref{preprocess}).

While they implemented early stopping, the use of a validation set is not mentioned in their study. Additionally, hyperparameters such as the learning-rate decay rate, momentum value, and L1 regularization constant are unspecified, which we fine-tune ourselves.

\textbf{Two-Stream Adaptive Graph Convolutional Networks (2s-AGCN)}~\cite{shi2019two} improves upon the STGCN by integrating dual stream predictions of bone vectors and joint coordinates. This model, leveraging a learnable graph structure and attention module, has demonstrated promising results in action recognition tasks. In our gait classification implementation, each stream of 2s-AGCN comprises three graph convolutional blocks with 64 input and output channels, followed by Global Average Pooling (GAP). Thereafter, the dual stream predictions are summed up to obtain the final model output.






 \setlength{\tabcolsep}{6pt}

\begingroup
\captionsetup{
  font=small, 
  labelfont=bf, 
  format=plain, 
  width=0.35\textwidth, 
}

\noindent
\begin{table}[tbp]
  \centering
  \begin{minipage}{.4\textwidth}
      \centering 
    \caption{FCN with early stopping on Walking gait in binary and multi-class settings. FCN shows a significant validation-test performance gap as gait categories(classes) increase. Models trained only with data of the specific classes used for comparison.}
  \label{tab:FCNonWG} 
 \begin{small}
  \begin{tabular}{@{} L{1cm} C{1cm} C{1cm} C{1cm} @{}}
 \toprule
\textbf{Class} & \textbf{Train Loss} & \textbf{Best Val Loss}& \textbf{Test Loss} \\   
    \midrule
    2 & 0.0003& 0.17 & 0.73 \\ 
    5 & 0.05 & 0.84 & 1.1 \\
    9 & 0.10 & 1.8 & 1.6  \\ 
     
    \bottomrule
  \end{tabular}
\end{small}
  \end{minipage}%
  \begin{minipage}{.58\textwidth}
  \captionsetup{
  font=small, 
  labelfont=bf, 
  format=plain, 
  width=1.01\textwidth, 
}

    \centering 
    \includegraphics[width=.9\linewidth]{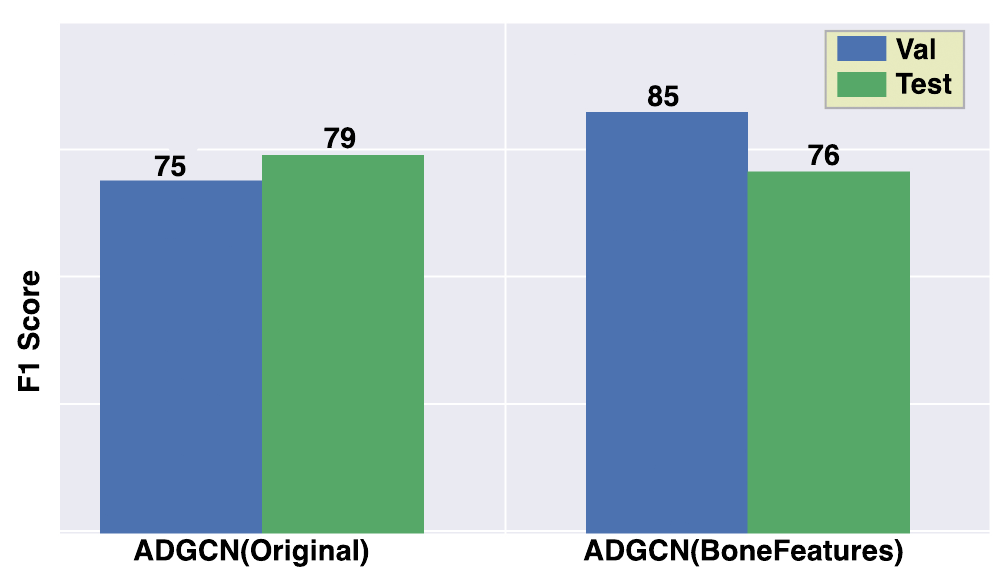}

     \captionof{figure}{ADGCN performance with and without bone features on PD-Walk dataset. Bone features adaptation adds a third branch of input, pairing bone length and angles locally. Test set performance did not improve on including bone features.}
       \label{tab:ADGCN_three} 


    \end{minipage}

\end{table}
\endgroup

\textbf{Multiple-Input Branch STGCN} \cite{kim2022pathological} uses multiple input branches, in contrast to multiple networks of 2s-AGCN. This model fuses joints, velocity, and bone inputs with three STGCN blocks and processes the concatenated outputs with two STGCN blocks. The two tested variants -  with and without attention - show equivalent performance. Consequently, we choose the simpler attentionless configuration for our experiments. Due to the lack of data split specifications, we adopt the same splitting mechanisms used in our other experiments (See Appendix Table \ref{tab:dataSplit}).
For missing hyperparameters, we draw insights from other STGCN-based modalities like \cite{song2020stronger} and conduct multiple tests with different hyperparameter combinations to report the best results.

\textbf{Asymmetric Dual-Stream Graph Convolution Network (ADGCN)}~\cite{he2022integrated} is a dual-stream GCN that utilizes joint coordinates, velocity, and acceleration, feeding these features in pairs to three STGCN blocks. The network employs two joint graphs: local connections consisting of natural joint connections and global connections where each joint connects to a central joint, a scheme also adopted in our proposed model (Figure \ref{fig:kinect25}). However, our model differs in two key aspects. First, while ADGCN combines inputs in pairs (trajectory + velocity and velocity + acceleration), we train models independently for each input. Second, we incorporate bone features in our model.

In the original paper, the ADGCN model is trained for 32 epochs with the Adam optimizer, a learning rate of 3e-4 with cosine annealing, and a weight decay of 5e-6. The results are reported using Focal loss and Cross Entropy loss, but Focal loss hyperparameters are not specified. We implement both losses and find the performance similar(see Figure ~\ref{fig:distributionShift}). Therefore, all our ensuing comparisons use Cross-Entropy loss to ensure a fair assessment against other models.

 \setlength{\tabcolsep}{6pt}

\subsection{Benchmarking Results}
\label{benchmarking_results}
As anticipated, our replication study of the aforementioned models reveals distribution shifts across different datasets when evaluated over different splits, which we quantitatively characterize to provide a holistic perspective on their overall performance (Figure ~\ref{fig:distributionShift}).


\begin{table*}[bp]
    \caption{F1 scores for FCN, 2sAGCN, MIB-STGCN and ADGCN on different datasets. We see a considerable val-test gap for most of the models. See Appendix Table \ref{tab:Hyperparameters}  for hyperparameter settings.*NC=No Convergence }
        \label{tab:benchmarking}
    \vskip 0.15in
    \begin{small}
 \begin{tabular*}{\linewidth}{@{\extracolsep{\fill}} lcccccccc }
    \toprule
    \multirow{2}{*}{Dataset} &
      \multicolumn{2}{c}{\textbf{FCN}} &
      \multicolumn{2}{c}{  \textbf{2sAGCN}} &
    \multicolumn{2}{c}{\textbf{MIB-STGCN}} &
        \multicolumn{2}{c}{\textbf{ADGCN}} \\
      & {Val} & {Test} & {Val} & {Test}  & {Val} & {Test}  & {Val} & {Test}  \\
      \midrule
    MMGS & 75.2 & 70.1  &  93.3 & 83.2 & 83.0 & 76.7 & 81.1 & 71.8  \\
    Walking Gait & 58.1 & 64.1 &  65.3 & 55.3 & 60.4 & 50.0 &*NC & *NC\\
    Pathological Gait & 88.2 &  85.5 & 98.0 & 95.6 & 93.4 & 88.1  & 96.4 & 93.9 \\
    PD-Walk & 77.1 & 77.2 &  84.1 & 83.1 & 80.1 & 80.1 & 75.1 & 79.1 \\
    \bottomrule
  \end{tabular*}
    \end{small}
\vskip -0.1in
\end{table*}

\textbf{Fully Convolutional Networks  (FCN)}~\cite{sadeghzadehyazdi2021modeling}.
Table \ref{tab:FCNonWG} presents loss values with our reimplementation on the Walking Gait dataset. As the number of classes increase, train-test performance gap escalates.
These results offer insights into the kind of misrepresentation that can occur without proper data splits.
Table \ref{tab:benchmarking} compares performance across all datasets.

\textbf{Two-Stream Adaptive Graph Convolutional Networks (2s-AGCN)}~\cite{shi2019two} was designed for action recognition tasks, incorporating nine Adaptive Graph Convolutional Network (AGCN) blocks. Our replicability study employs three AGCN blocks with 64 output channels to prevent overfitting on gait data. Its results across all datasets (Table \ref{tab:benchmarking}) offer acceptable threshold values for comparing with our proposed model.





\textbf{Multiple-Input Branch STGCN (MIB-STGCN)}~\cite{kim2022pathological}.
This model uses five STGCN blocks. 
Testing on multiple datasets (Table \ref{tab:benchmarking}) reveals significant variations between validation and test accuracies for this model.

\textbf{Asymmetric Dual-Stream Graph Convolution Network (ADGCN)}~\cite{he2022integrated}. 
The original ADGCN network, as tested with early stopping in our experiments, converged within the initial ten epochs, and further training only lead to overfitting. Table \ref{tab:benchmarking} presents its results across all datasets. Notably, the model failed to converge on the Walking Gait dataset, resulting in Validation and Test F1 scores below 40\%.

We also tested an ADGCN modification that utilizes the same input features as our proposed model. Figure \ref{tab:ADGCN_three} compares our adaptation—integrating bone features (local graph with bone length and bone angles) as the third input branch and the original ADGCN architecture. While including bone features improved performance on the validation set, this enhancement did not consistently extend to the test set. This suggests that utilizing paired features does not consistently guarantee superior performance. In contrast, our proposed model, involving independent experts, demonstrates consistent improvements, as depicted in Table \ref{tab:reswithAMSGCN}.

  
      




In conclusion, our findings suggest that despite the prolific use of STGCN variants in gait classification, their implementation procedure is often unreliable. Our study indicates that these models tend to overfit to the details of a specific training subset, resulting in poor generalization on the test set. While 2s-AGCN demonstrated acceptable performance, some modifications exhibited a test-time performance deterioration. These observations emphasize the need for careful validation when employing STGCN variants for gait classification tasks.

\section{Asynchronous Multi-Stream Graph Convolutional Network (AMS-GCN)}
\label{amsgc_intro}
In high-dimensional settings, deep learning models often converge poorly, showing promising results only on specific datasets or on certain subspaces of the dataset. The stochastic nature of these optimizations can lead to overfitting on training data, sabotaging performance on the test data. An ensemble can mitigate such problems by averaging over different local optima, preventing any single model from dominating the predictive performance \citep{sagi2018ensemble}.

Ensemble learning, characterized by expert committees, involves training these experts on specific input feature spaces and amalgamating their expertise to enhance generalization. Drawing inspiration from \cite{shi2019two}, we introduce a Multi-Stream network that incorporates global and local branches, exposing distinct pathologically relevant features for different disorders. 
Our model uses five features - joint coordinate, velocity, acceleration, bone vector, and bone angle (Appendix Table 11). For each feature, we have two networks  - one local network that exploits the natural human-body joint relations, and one global network where all joints are connected to the "SpineBase" joint, one of the twenty-five joints provided by the Kinect (Figure \ref{fig:kinect25}). 
Thus, we have a total of ten networks in the ensemble. 

The ten networks can be trained independently and parallelly for a classification task. 
Each of these consists of three STGCN blocks with each block having 64 output channels followed by a global average pooling layer. Finally, we apply a fully connected layer with 64 inputs.
We use Cross Entropy loss with train-test-validation splits and early stopping for training each network.

To get the final ensemble prediction, we first add the raw output from a feature's local and global network and convert this sum into probability scores using the softmax function. These values form the prediction score from a single feature. Finally, we average the probability scores from the five features to get the desired AMSGCN prediction.
Please see Appendix Section \ref{AMSGCNarchitecture} for preprocessing details.





\section{AMS-GCN Classification Analysis}
\label{amsgcn_analysis}

To investigate our model's performance, we train AMS-GCN over multiple train, validation, and test splits on all the datasets used in this paper. AMS-GCN performs well on all the datasets except for Walking Gait (Table~\ref{tab:reswithAMSGCN}). As we describe below, Walking Gait is hard to classify because of subtle inter-class variations. For AMS-GCN, validation and test scores are consistent across all other datasets, exhibiting the model's robust generalization capabilities.
See Appendix Section \ref{threshold_agnostic} for threshold agnostic comparisons.

\subsection{Multi-Modal Gait Symmetry (MMGS) Dataset}
\citet{khokhlova2019normal} used proper train-validation-test splits for assessing this dataset, which we reuse to maintain consistent performance validation. Their findings reveal significant variance between validation and test scores across different splits. This can be due to inherent noise present during data capturing. Different data splits could also be capturing different demographic categories as this dataset contains features from 19 men and 8 women aged 23 to 55 (standard deviation of seven years). 
Notably, AMS-GCN outperforms all competitors, and demonstrates consistent performance between the validation and test sets (Table \ref{tab:reswithAMSGCN}).

\subsection{Walking Gait Dataset}
In their study, \citet{nguyen2016skeleton} used five out of nine subjects for training and the remaining four for testing. In our approach, we use five subjects for training, allocate two for validation, and reserve two for testing. Notably, diagnosing treadmill walking proves to be more challenging than regular walking. As indicated in Table \ref{tab:reswithAMSGCN}, all models exhibit suboptimal performance. 
A more granular analysis shows that classes 4 and 8 (15cm padding sole under left and right feet) are difficult to separate for all the experts of AMS-GCN (Table~\ref{tab:ClassresOnWalkingGait}). We also notice that different input features capture class-specific properties, endorsing our hypothesis of an ensemble approach. These findings also suggest that a weighted average ensemble model may be more suitable for this dataset.

 \setlength{\tabcolsep}{3pt}

\begin{table*}[tbp]
    \caption{Validation and Test time weighted-average F1 scores for all models on all the datasets. All PD-Walk results are for 10 runs over two data splits, all other results are for 15 runs over three data splits. AMS-GCN performs better than all other models both in terms of F1 scores and test-validation gap. *ADGCN did not converge on Waking Gait, and plateaued at around 40\% F1 score,NC = No Convergence.}
        \label{tab:reswithAMSGCN}
    \begin{small}
  \begin{tabular*}{\linewidth}{@{\extracolsep{\fill}} lcccccccc }
    \toprule
    \multirow{2}{*}{Model} &
      \multicolumn{2}{c}{\textbf{MMGS}} &
      \multicolumn{2}{c}{  \textbf{Walking Gait}} &
    \multicolumn{2}{c}{\textbf{Pathological Gait}} &
      \multicolumn{2}{c}{\textbf{PD-Walk}} \\
      & {Val} & {Test} & {Val} & {Test}  & {Val} & {Test} & {Val} & {Test}  \\
      \midrule
    FCN & 75.2$\pm$8.1 & 70.1$\pm$6.7 &  58.1$\pm$2.8 & 64.1$\pm$1.4 & 88.2$\pm$7.0 & 85.5$\pm$2.1 & 77.1$\pm$5.7 & 77.2$\pm$3.2 \\
    MIB-STGCN & 83.0$\pm$5.5 & 76.7$\pm$6.8 &  60.4$\pm$8.8 &50.0$\pm$7.2 & 93.4$\pm$3.1 & 88.1$\pm$3.4 & 80.1$\pm$4.1 & 80.1$\pm$4.0  \\
    2s-AGCN & 93.3$\pm$4.6 & 83.2$\pm$2.9 &  65.3$\pm$4.9 & 55.3$\pm$6.5 & 98.0$\pm$1.3 & 95.6$\pm$3.0 & 84.1$\pm$4.2 &83.1$\pm$4.8 \\
        ADGCN & 81.1$\pm$8.8 & 71.8$\pm$10.7 &  NC* & NC* & 96.4$\pm$2.7 & 93.9$\pm$3.4 & 75.1$\pm$5.0 & 79.1$\pm$6.0\\
        AMS-GCN & 91.2$\pm$3.9 &\textbf{88.3$\pm$3.7} &  73.4$\pm$3.4 & \textbf{65.9$\pm$4.5} & 98.6$\pm$1.3 &\textbf{ 96.1$\pm$2.1} & 84.8$\pm$1.5 & \textbf{85.4$\pm$2.1} \\
    \bottomrule
  \end{tabular*}
    \end{small}
\end{table*}

\subsection{Pathological Gait Dataset}
The dataset paper \citep{jun2020pathological} utilizes leave one subject out cross-validation for reporting results.
It has data from 10 subjects, so we fixed six subjects for training and two each for validation and testing. Table \ref{tab:reswithAMSGCN} clearly shows that AMS-GCN outperforms all other models. Other models also perform comparatively well on this dataset. One reason for this could be the use of multiple Kinects, which facilitates robust data collection. Deep learning algorithms are vulnerable to outliers. Hence, a less noisy dataset would generally procure higher accuracy.

\subsection{PD-Walk - real Parkinson's Patient Dataset}
The dataset paper offers details on multiple train-test splits but does not include a validation set. To address this, we split their original test set into 40\% validation and 60\% test data. All models are steady between validation and test results (Table \ref{tab:reswithAMSGCN}). Our proposed model beats the State of the Art algorithm (ADGCN from PD-Walk paper). 
The success of machine learning with a real Parkinsonian gait dataset demonstrates its potential to aid geriatricians in identifying early signs of neurodegenerative disorders. 


\begin{table*}[tbp]
  \centering 
      \caption{Class-specific F1 scores for five expert models of AMS-GCN on the test set of Walking Gait dataset for 9 classes. Test subjects[2,7], Class 1 is normal walk. Certain features show good scores on specific gait classes, advocating a weighted ensemble approach for this dataset.}
          \label{tab:ClassresOnWalkingGait} 

    \begin{small}
  \begin{tabular}{lccccccccc}
  \toprule
    \textbf{joint input} & \textbf{1} & \textbf{2} & \textbf{3} & \textbf{4} & \textbf{5} & \textbf{6} & \textbf{7} & \textbf{8} & \textbf{9}  \\
    \midrule
    coordinates &  0.59  &   0.64  &   0.46  &   0.52  &   0.66  &   \textbf{0.78}  &   0.52  &   0.41  &     0.80 \\ 
    velocity &  0.60 &   0.44 &   0.38 &   0.37 &   \textbf{0.79} &   0.38 &   0.14 &   0.49 &   0.52 \\ 
    acceleration &  0.42 &    0.30 &    0.34 &    0.48 &    0.26 &    0.41 &    0.01 &    0.50 &    0.73 \\
    bone vectors &  0.31 &   0.31 &   0.25 &   0.00 &   0.65 &   0.50 &   0.57 &   0.52 &   \textbf{0.83} \\
    bone angles &   0.40  &    \textbf{0.79}  &    0.57  &    0.46  &    0.80  &    0.46  &    0.42  &    0.39  &    0.54 \\
    \bottomrule
  \end{tabular}

      \end{small}

\end{table*}

\section{AMS-GCN: Performance Insights}
\label{amsgcn_insights}

\subsection{Ensembling Techniques}
\label{ablation_ensemble}

   
     

To analyze how the ten expert committees of AMS-GCN supplement each other, we test various ensembling techniques such as \textbf{Hard Weighted Majority Voting, Soft Weighted Majority Voting, and Stacking}. For this analysis, we utilize the MMGS dataset ~\citep{khokhlova2019normal} and report the aggregate results of ten runs over two data splits.

\textbf{Weighted Majority Voting Algorithm} - In this voting scheme, for every wrong ensemble prediction we update the weights of all the incorrect models according to the following scheme:
\[ w_{t+1} = w_t*beta, \;\;\; \beta=0.5 \]
All the models start with an initial weight of one.
In the hard voting case, the weights of the models having the same prediction are summed to form the ensemble prediction, whereas, in the soft voting case, individual model weights and predictions are multiplied and added to generate the ensemble prediction. The weights are updated using the validation set. 

In \textbf{Stacking}, the validation set experts' predictions are used to train another classifier, which predicts the final class for a sample. During test time, this classifier takes the committees' outputs and predicts the test sample label. For our comparisons, we used the XGBoost classifier.
We also tested a \textbf{Joint Training} model by adding another feedforward layer on top of the last STGCN block (having sixty-four-bit output) of each expert. This ensemble trains all models together and gives one consolidated prediction for the ten experts. However, the results of this setup were not so encouraging. 



Moreover, the other tested ensembles also could not come close to the performance of our AMS-GCN
(Table \ref{tab:ensemble}). We hypothesize that even for the same gait abnormality, different feature combinations dominate classification on different data splits, just as the same abnormality can be simulated using different techniques. Hence, relying heavily on certain features grounded on an isolated data segment deteriorates the performance on other segments. This is probably why many of the benchmarked methods do not work when tested across different data splits. They learn specialized feature combinations not universally relevant for all data splits, which is taken care of by an equi-weighted ensemble model.


   
     
  

\begingroup
\captionsetup{
  font=small, 
  labelfont=bf, 
  format=plain, 
  width=1\textwidth, 
}

\noindent
\begin{table}[tbp]
  \centering
  \begin{minipage}{.45\textwidth}

      \centering 
    \caption{Ensemble methods comparison using test time F1 scores on MMGS dataset over ten runs with two data splits.}
  \label{tab:ensemble} 
\begin{small}
 \begin{tabular}{@{} L{4cm} C{1.5cm}  @{}}  
 \toprule
  \textbf{Method}  & \textbf{Test F1} \\
   
    \midrule
  Hard Weighted Majority    & 79.2$ \pm $6.6  \\ 
  Soft Weighted Majority   & 80.4$ \pm $3.0 \\
  Stacking                  & 82.7$ \pm $3.4 \\ 
  Joint Training            &  76.6$ \pm 6.6 $\\
  AMS-GCN                   & \textbf{88.3$ \pm $3.0} \\
     
    \bottomrule
  \end{tabular}
\end{small}
  \end{minipage}%


\end{table}

\endgroup

 \setlength{\tabcolsep}{6pt}
 
\begin{table*}[bt]
  \centering 
   \caption{AMS-GCN Input feature comparison - best and worst feature combinations and respective F1 scores. Vel=Velocity, Acc=Acceleration, boneL = bone Length and boneA = bone Angle. Whenever two models have the same F1 score, we report one of them randomly. The "Method" column shows the number of features combined to test the ensemble methods. Pathological Gait dataset is not compared as it is relatively easy to classify.}
  \label{tab:ablation_feat} 

   \vskip 0.15in
\begin{small}
  \begin{tabular}{lll}
  \toprule
  \textbf{Method}  & \textbf{Worst F1}  & \textbf{best F1} \\
    \midrule
  \multicolumn{3}{c}{MMGS}  \\
  1 & Acc(56.0)  & boneA(86.0) \\
   2 & Vel+Acc(55.0)      & boneL+boneA(87.0) \\ 
  3 & Joint+Vel+Acc(77.0) & Acc+boneL+boneA(87.0) \\
   4 & Joint+vel+Acc+boneL(82.0)      & Joint+Acc+boneL+boneA(86.0) \\ 
  \multicolumn{3}{c}{PD-Walk}  \\
    1 & Acc(17.0)  & boneL(87.0) \\
   2 & Vel+Acc(17.0)      & boneL+boneA(86.0)\\ 
  3 & Vel+Acc+boneA(83.0)  & Acc+boneL+boneA(86.0) \\
   4 & Joint+Vel+Acc+boneA(85.0)      & Vel+Acc+boneL+boneA(86.0) \\ 
  \multicolumn{3}{c}{Walking Gait}  \\
    1 & Joint(41.0)  & Acc(62.0) \\
   2 & Vel+Acc(54.0)      & Joint+Acc(66.0)\\ 
  3 & Joint+boneL+boneA(55.0)  & Vel+Acc+boneA(71.0) \\
   4 & Joint+Vel+Acc+boneL(63.0)      & Joint+Acc+boneL+boneA(64.0) \\ 
     
    \bottomrule
  \end{tabular}
 \end{small}
\vskip -0.1in
\end{table*}

\subsection{Feature Utility.}
\label{ablation_feature}

To gauge the effect of different gait features on classification score,
we run AMS-GCN with different combinations of features, using single, pair, triplets, and four features at a time. Table \ref{tab:ablation_feat} reports the best and worst-performing combinations for each set over ten runs on two data splits of each dataset.

The results show that different datasets have different (combinations of) dominating features. For example, as a single feature, Acceleration does worst on MMGS \citep{khokhlova2019normal}, whereas it does best on Walking Gait~\citep{nguyen2016skeleton}. We also find that bone features are always part of the best-performing model and thus play a crucial role in differentiating various pathological conditions. However, as we move towards using four or five features, all the models show almost equivalent performance between best and worst feature combinations.

Therefore, from single and pair comparisons we gain insights into the significance of certain features over others. Notably, among comparably performing combinations, removing bone angles almost always degrades results. Similarly, joint coordinates also show up in many of the best-performing combinations. Velocity and acceleration in isolation show bad performance but could be subtly affecting the overall ensemble score. Furthermore, as we increase the number of features in a direct combination, the performance fluctuations settle down.
We anticipate that this hierarchical exploration of gait features will help catalyze future advancements in gait assessment methods.


\subsection{Performance with Benchmarked Model Inputs}
For a more direct comparison against other models, we narrow down the AMS-GCN input space to match that of respective models, as detailed in Table \ref{tab:AblationFeatAMSGCN}, and compare them over multiple runs on multiple splits. The results indicate that even when using a limited number of features, a direct local-global feature ensemble is most robust against subtle variations in gait anomalies. 
After conducting various tests on the Walking Gait dataset, we found FCN showing high loss values albeit better test accuracy. We hypothesize that FCN might be memorizing patterns in the training set, resulting in a significant loss on specific examples from validation and test sets.
Notably, AMS-GCN on Walking Gait with reduced ADGCN features set performed better than using all five features, advocating for a weighted ensemble for this dataset.

\textit{Please check \href{https://github.com/jlabhishek/AMSGCN-MLHC-24}{github} for the code.}

\begin{table*}[bt]
\centering
    \caption{Head-to-head F1 score comparison of AMS-GCN with different models using the same feature set incorporated in compared models. Refer Appendix Table \ref{tab:DatasetFeatures} for the specific feature set used by each model. AMS-GCN performs better than other models even with a reduced feature set. The closest competitor is 2s-AGCN, which does not use global graphs. All results are averaged over 10-15 runs over three data splits.*NC=No Convergence }
        \label{tab:AblationFeatAMSGCN}
\begin{small}
  \begin{tabular*}{\linewidth}{@{\extracolsep{\fill}} lcccccccc }
    \toprule
    \multirow{2}{*}{Dataset} &
          \multicolumn{2}{c}{  \multirow{2}{*}{
          \parbox{2cm}{\centering FCN\\Features} }} &

      \multicolumn{2}{c}{\multirow{2}{*}{\parbox{3.2cm}{\hspace{.25cm } \centering 
 MIB-STGCN\\\hspace{.2cm }Features} }} &

    \multicolumn{2}{c}{\multirow{2}{*}{\parbox{2.2cm}{\centering 2s-AGCN\\Features} }} &

      \multicolumn{2}{c}{\multirow{2}{*}{ \parbox{2.2cm}{\centering ADGCN\\Features} }} \\
     & & & & & & & &\\
      & {FCN} & {Ours} & {MIB-STGCN} & {Ours}  & {2s-AGCN} & {Ours} & {ADGCN} & {Ours}  \\
      \midrule
    MMGS & 70.1 & \textbf{81.4} &  76.7 & \textbf{85.6} & 83.2  &\textbf{84.8} & 71.8 & \textbf{79.1} \\
    Walking Gait & \textbf{64.1} & 61.8 &  50.0 & \textbf{62.1} & 55.3 & \textbf{55.5} & *NC & \textbf{67.2} \\
   Pathological Gait & 85.5& \textbf{93.3} &  88.1 &\textbf{96.0}  & 95.6 & \textbf{95.9} & 93.9 & \textbf{94.5}\\
        PD-Walk & 77.2 & \textbf{80.3} &  80.1 & \textbf{84.2} & 83.1 & \textbf{83.9} & 79.1 & \textbf{81.6}\\
       
    \bottomrule
  \end{tabular*}

      \end{small}
\vskip -0.1in
\end{table*}

\section{Discussion} 


Previous studies~\citep{sadeghzadehyazdi2021modeling,kim2022pathological,tian2022skeleton} have incorporated features such as joint coordinates, motion signals at different scales, and joint adjacency to guide the classification process. However, none have articulated results that combine all such features, especially with local and global graph structures. Although ~\cite{he2022integrated}, use local and global graphs, as described earlier in Section \ref{benchmarking_results}, a direct combination like ours performs better than their paired inputs. 
In addition, after experimenting with various ensembling methods, our tests reveal that, in such high-dimensional settings, a straightforward combination works best. This finding motivates the final ensembling technique chosen by us. Thus, our AMS-GCN model, even though superficially a simple ensemble, distinguishes itself from previous works.
After conducting comparisons across four datasets with multiple runs and splits, our results suggest that the proposed AMS-GCN model establishes itself as a robust baseline for future endeavors and a viable alternative to previous research efforts.
In summary, the contributions of this paper are threefold. First, we benchmark the performance of algorithms developed for gait classification on four different datasets, testing their generalized ability to identify various pathological gait conditions. A critical reporting mistake seen in this literature is not using separate validation and test sets, thus leaving open the possibility of over-fitting to the test set during training~\citep{dwork2015generalization}. Another fallacy is the common practice of tuning model hyperparameters directly on the test
set \citep{recht2019imagenet}. We report results using a test-validation split, and find distribution shifts of between 10-15\% (Figure \ref{fig:distributionShift}) across datasets for most models, in line with recent generalization results in related settings~\citep{recht2019imagenet}. 

\begin{figure}[tbp]
\centering
\includegraphics[width=0.6\linewidth]{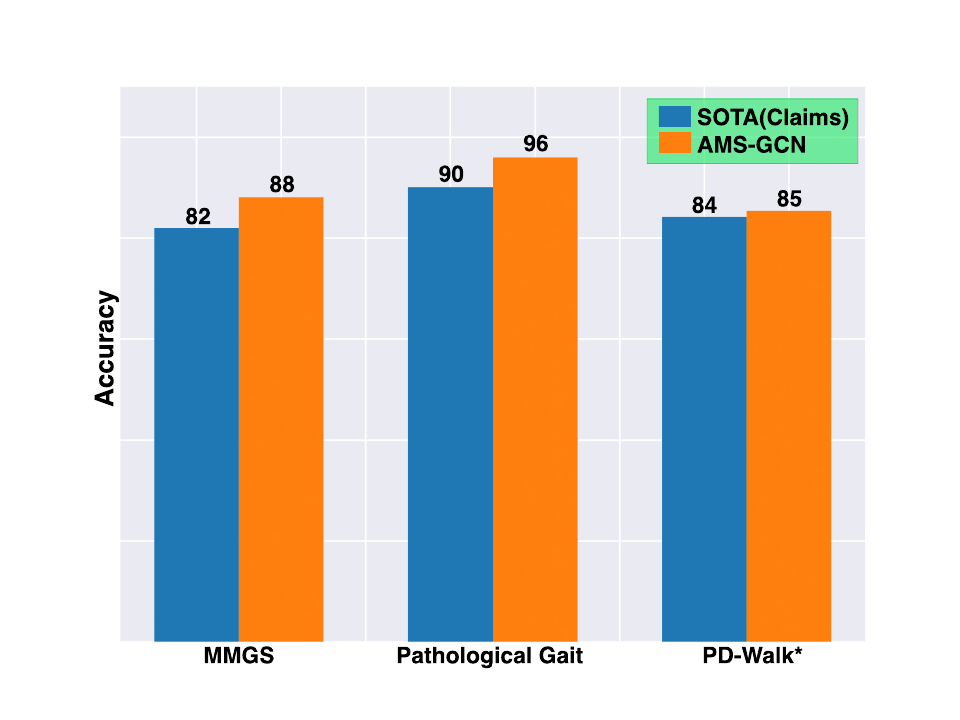}
\vskip -.2in
\caption{Accuracy comparison for AMS-GCN against the best previously reported results. All AMS-GCN results are averaged over multiple data splits. *F1 score using Cross Entropy Loss. For Walking Gait, all models perform poorly. }
\label{fig:SOTA}
\end{figure}

Second, we present a robust model tested on multiple datasets that demonstrates good generalizable performance to support clinicians for early diagnosis in home settings, outperforming state-of-the-art models by between 5-10\% across datasets in most cases (Figure~\ref{fig:SOTA}).

Finally, through various tests employing different feature combinations, we capture insights into the efficacy of different feature combinations for gait assessment. 
This foundation paves the way for future research, particularly in developing clinical studies aimed at further investigating these crucial gait features, thereby bolstering efforts in automatic gait classification.


Our replication study revealed a significant amount of distribution shift for existing studies. We also showed that in complex optimization scenarios, models are susceptible to showing promising results on one part of the dataset while doing poorly on other parts, supporting the reporting of disaggregated error analyses in this area of machine learning~\citep{burnell2023rethink}. Previously, \citet{zhang2021understanding} have shown that large neural networks have enough capacity to perfectly shatter the training set, fitting a random labeling of the training data and also true images with noise. Along the same lines, we found that STGCN-based models are prone to overfitting on typically sized datasets in this area.
To actually realize the potential of a proposed algorithm, experiments must be conducted over 
multiple data splits, requiring algorithms to consistently perform well across all splits.

We also strongly recommend using early stopping in training as a safeguard against fitting noise in the data, especially when working with pose estimation methods where limbs can be easily deformed across video frames. 
Ensuring reliability is crucial for the practical application of healthcare technologies. Therefore, given the intricate nature of gait data, adopting a streamlined AMS-GCN ensemble approach that treats all models uniformly proves to be advantageous.

\paragraph{Limitations}
While we report heterogeneities in model performance across data splits in several of our experiments, we ourselves report only aggregate measures of performance for ease of comparison with the existing literature. Future work should focus on reporting disaggregated evaluations on 'hard' data~\citep{burnell2023rethink}.

Many other variants of STGCN models in the action recognition field could work even better for pathological gait classification, which needs further exploration. Our proposed method achieves good performance because a mixture of experts shades individual models' inadequacies, making up for occasional poor convergence. 

In this study, we did not evaluate the efficacy of using different sets of joints for classification, which needs further exploration and can be examined in an extension of the current work.

\bibliography{ref}

\newpage
\appendix

\section{Datasets - Feature Selection and Preprocessing}
\label{preprocess}
We follow standard and identical preprocessing pipelines for all Kinect-based datasets. Each Kinect dataset captures the x, y, and z coordinate time series for 25 joints. First, we identify V relevant joints for differentiating pathological gaits and subtract the per-frame "SpineMid" joint (Figure \ref{fig:kinect25}) coordinates from each joint across all frames. While \citet{tian2022skeleton} advocate using selected lower body joints, we maintain a more general setting to facilitate meaningful comparisons.

Then, we split these normal and simulated walks into multiple data clips of length T, augmenting them by overlapping 40 frames between successive clips. The Pathological Gait dataset, being sufficiently large, is exempt from such augmentation. 

Finally, for each frame, we range normalize each x, y, and z input by applying the following formula on each coordinate of the selected V joints. 
\begin{equation} k_{normalized} = \frac{k - k_{min} } {k_{max} - k_{min}} \end{equation}
where k $\epsilon$ V and $k_{min}$ and $k_{max}$ are the min and max value of k for each data clip i.e., across T frames. This gives us normalized model input for T frames and V joints. We use T = 48 and V = 19 for all Kinect-based experiments. Consequently, we use 19 of the 25 Kinect body tracking joints removing the fingers, "SpineMid", and Head joints.

PD-Walk involves different sets of joints than Kinect. Here, we exclude head information and use the remaining 12 joints (=V) for further analysis (refer to Table \ref{tab:dataSplit} for a description of the train, validation, and test splits used in investigating the existing models).

\section{AMS-GCN Model Architecture and Preprocessing}
\label{AMSGCNarchitecture}
We use the following procedure to normalize the 3D Joints coordinates for the V selected joints  (V=19 for Kinect-based datasets and 12 for PD-Walk).

   \begin{equation}v_t = x_{t+1} - x_{t}   \end{equation}
  \begin{equation} a_t = v_{t+1} - v_{t}  \end{equation}
    \begin{equation} b_t = x - x_{adj}   \end{equation}
    \begin{equation} b_a = \arccos( \frac{b_t}{||b_t||}) \end{equation}

where $x_t$, $v_t$, and $a_t$ are coordinates, velocity, and acceleration at time t for each joint. $b_t$ is the bone vector defined as the difference in coordinate values of adjacent joints according to the graph connections. $b_a$ is the bone angle corresponding to each bone vector.
We also add a residual connection in our STGCN blocks such that the input of each block is added to its output before applying the activation function and transmitting information to the next block.

\section{Hyperparameters and Data splits}
We train all the models (except FCN) using Adam optimizer and early stopping on validation loss with batch size 32. The base learning rate was 0.1, with a weight decay rate of 0.001. We train the models for 200 epochs and update the learning rate after every 80 epochs as one-tenth of the previous value. FCN \citep{sadeghzadehyazdi2021modeling} uses L1 regularization with regularization constant 1e-5 and is trained for 400 epochs with early stopping.

We use the following settings for training all the experts of AMS-GCN :
Epochs = 200, Optimizer = Adam, Batch size = 32, weight decay = 1e-3, base learning rate = 0.1, Learning rate decay = 0.1 * learning rate at epochs 80, 160, and 180. Early stopping on validation loss.

Table \ref{tab:DatasetClassFrequency} shows the per-class frequency for all the datasets and Table \ref{tab:dataSplit} shows data splits for all the datasets used in our experiments.



\begin{table*}[tbhp]
  \centering 
  \caption{Class frequency for each class for all datasets. Each class is split into Train/Val/Test across multiple splits for results comparison. Class 1 is the correct class. *Bracketed number denotes the total number of classes in the dataset}.
      \label{tab:DatasetClassFrequency} 
\vskip 0.15in
\begin{small}
  \begin{tabular}{lllllll}
  \toprule
    \textbf{Dataset}  & \textbf{Class 1} & \textbf{Class 2} & \textbf{Class 3} & \textbf{Class 4} & \textbf{Class 5} & \textbf{Class 6}  \\
    \midrule
    MMGS(3)*    &  1221 & 1220 & 1406 & - & - & -\\ 
    Walking Gait(9)*  & \multicolumn{6}{c}{1504 for each of the 9 classes} \\
    Pathological Gait(6)*   & 1199 &1186 & 1197 & 1196 & 1196 & 939 \\ 
    PD-Walk(2)*  & 4482 & 5390 & - & - & - & -  \\ 
    \bottomrule
  \end{tabular}
  \end{small}
\vskip -0.1in

\end{table*}

\begin{table*}[tbhp]
  \centering 
   \caption{Subject Wise data splits for different Pathological Gait datasets reusing the splits used in the dataset papers.}
  \label{tab:dataSplit} 

   \vskip 0.15in
\begin{small}
  \begin{tabular}{llll}
    \textbf{Splits} & \textbf{Train} & \textbf{Val } & \textbf{Test }  \\
      \toprule
    \multicolumn{4}{c}{MMGS(22 subjects)} \\
    Split 1 & [2,3,7,8,9,11,13,14,15,19,21] & [1,6,10,17,18] & [4,5,12,16,20,22] \\ 
    Split 2 & [2,5,8,10,11,13,14,15,16,19,20] & [1,4,17, 18,21] & [3,6,7,9,12,22] \\
    Split 3 & [2,3,8,10,12,13,16,17] & [5,6, 9,11,21]  & [1,4,7,14,15,18,19,20,22,] \\ 
    \multicolumn{4}{c}{Walking Gait } \\
    Split 1 & [1,3,5,6,9]  & [4,8] & [2,7]  \\
    Split 2 & [1,3,5,6,9]  & [2,7] & [4,8]  \\
    Split 3 & [1,3,5,6,9]  & [2,4] & [7,8]  \\
        \multicolumn{4}{c}{Pathological Gait } \\

    Split 1 & [1,2,4,7,8,10]  & [3,6] & [5,9]   \\ 
    Split 2 & [1,2,6,7,9,10]  & [3,5] & [4,8]   \\ 
    Split 3 & [1,3,5,7,9,10]  & [4,8] & [2,6]   \\ 

             \multicolumn{4}{c}{PD-Walk  } \\
    Split 1 &  Split 1 of dataset paper & 40\% of test data & 60\% of test data    \\ 
    Split 2 &  Split 3 of dataset paper & 40\% of test data & 60\% of test data    \\ 
  \end{tabular}
 \end{small}
\vskip -0.1in
\end{table*}

\begin{table}[tbhp]
  \centering 
  \caption{Hyperparameters used for training various models in the benchmarking study. LR=Learning Rate, Ep=Training Epochs, L1 const. is the L1 regularisation constant.}
        \label{tab:Hyperparameters} 

    \vskip -0.1in
    \begin{small}
  \begin{tabular}{llll}
  \toprule
     \textbf{FCN} & \textbf{MIB-STGCN} & \textbf{2sAGCN}  &\textbf{ADGCN}\\
    \midrule
     \multirow{4}{*}{\parbox{2.5cm}{\raggedright LR=0.1, Ep=400, LR=0.1*LR every 100 ep, L1 const.=1e-5}}& \multirow{4}{*}{\parbox{2.5cm}{\raggedright LR=0.1, Ep=200, Wt.Decay=0.001, LR=0.1*LR every 80 epochs}} & \multirow{4}{*}{\parbox{2.5cm}{\raggedright LR=0.1, Ep=200, Wt. Decay=0.001, LR=0.1*LR every 80 epochs.}}  &\multirow{4}{*}{\parbox{2.5cm}{\raggedright LR=0.1, Ep=200, Wt. Decay=.001, LR=0.1*LR every 80 epochs.}} \\ 
     & & & \\
        & & & \\
     & & & \\
     & & & \\

    \bottomrule
  \end{tabular}
\end{small}
\end{table}

\begin{table}[tbhp]
    \centering 
     \caption{Input features used by different methods compared in this study.}
       \label{tab:DatasetFeatures} 
      \vskip -0.05in
    \begin{small}
  \begin{tabular}{@{} L{2.2cm}  L{5cm} @{}}
  \toprule
  \textbf{Model}  & \textbf{Feature Set} \\
    \midrule
\textit{FCN} & Joint \\
\textit{MIB-STGCN} & \multirow{2}{4.5cm}{Joint + Velocity + Bone Vector + Bone Angles}  \\
\\
\textit{2s-AGCN} &Joint + Bone Vector \\
\textit{ADGCN} & Joint + Velocity + Acceleration \\
\textit{AMS-GCN} & \multirow{2}{5cm}{Joint + Velocity + Acceleration + Bone Vector + Bone Angles} \\
\\   
    \bottomrule
  \end{tabular}
      \end{small}

\end{table}
\section{Threshold Agnostic Metrics Comparisons}
\label{threshold_agnostic}
To further test AMSGCN, we also compare AUROC/AUPRC scores for all the models (Table \ref{tab:gait_analysis_auroc} and \ref{tab:gait_analysis_auprc}). These results further highlight the performance benefit of using AMSGCN. We also see that AMSGCN has better AUPRC values than other models on the Walking Gait dataset which is the most difficult dataset to classify among all tested.
\begin{table}[tbh]
    \centering
        \caption{AUROC (Mean + Standard Deviation) scores on pathological gait datasets.}

    \begin{tabular}{lcccc}
        \toprule
        \textbf{Model} & \textbf{MMGS} & \textbf{Walking Gait} & \textbf{Pathological Gait} & \textbf{PD-Walk} \\
        \midrule
        FCN       & 0.92±0.04 & 0.93±0.01 & 0.98±0.02 & 0.79±0.06 \\
        MIB-STGCN & 0.90±0.03 & 0.92±0.02 & 0.99±0.01 & 0.91±0.02 \\
        2sAGCN    & 0.94±0.02 & 0.94±0.01 & 0.99±0.01 & 0.91±0.06 \\
        ADGCN     & 0.87±0.06 & 0.82±0.01 & 0.99±0.01 & 0.90±0.03 \\
        AMSGCN    & \textbf{0.96±0.03} & \textbf{0.97±0.01} &\textbf{ 0.99±0.01} & \textbf{0.94±0.01} \\
        \bottomrule
    \end{tabular}
    \label{tab:gait_analysis_auroc}
\end{table}
\begin{table}[H]
    \centering
    \vskip -.1in
        \caption{AUPRC (Mean + Standard Deviation) scores on pathological gait datasets.}
    \begin{tabular}{lcccc}
        \toprule
        \textbf{Model} & \textbf{MMGS} & \textbf{Walking Gait} & \textbf{Pathological Gait} & \textbf{PD-Walk} \\
        \midrule
        FCN       & 0.86±0.06 & 0.59±0.01 & 0.95±0.04 & 0.84±0.01 \\
        MIB-STGCN & 0.82±0.04 & 0.66±0.06 & 0.95±0.02 & 0.95±0.01 \\
        2sAGCN    & 0.90±0.03 & 0.70±0.04 & 0.99±0.01 & 0.92±0.6  \\
        ADGCN     & 0.78±0.09 & 0.46±0.04 & 0.98±0.01 & 0.93±0.03 \\
        AMSGCN    &\textbf{ 0.93±0.03} & \textbf{0.77±0.03} & \textbf{0.99±0.01} & \textbf{0.96±0.01 }\\
        \bottomrule
    \end{tabular}

    \label{tab:gait_analysis_auprc}
\end{table}

\end{document}